\documentclass[letterpaper, 10 pt, conference]{ieeeconf}  

\IEEEoverridecommandlockouts                              

\usepackage{flushend}
\usepackage{graphicx}
\usepackage{mathptmx}
\usepackage{amsmath} 
\usepackage{amssymb}  
\usepackage{multirow}
\usepackage{threeparttable}
\usepackage[ruled]{algorithm2e}

\title{\LARGE \bf
6-DoF Grasp Detection in Clutter with Enhanced Receptive Field and Graspable Balance Sampling 
}

\author{Hanwen Wang$^{1}$, Ying Zhang*$^{1}$, Yunlong Wang$^{2}$ and Jian Li$^{1}$
\thanks{*Corresponding author.}
\thanks{$^{1}$Hanwen Wang, Ying Zhang and Jian Li are with 
     the  School of Intelligent Engineering and Automation, 
     Beijing University of Posts and Telecommunications, 
     Beijing, 100876, China.
        {\tt\small \{whw2022111391,
        yingzhang\_bupt\}@bupt.edu.cn}, {\tt\small jianli\_628@126.com}} %
\thanks{$^{2}$Yunlong Wang is with the State Key Laboratory of
Multimodal Artificial Intelligence Systems, Institute of Automation, Chinese Academy of Sciences, Beijing, 100190, china. 
 {\tt\small yunlong.wang@cripac.ia.ac.cn}} }

\begin{document}

\maketitle
\thispagestyle{empty}
\pagestyle{empty}

\begin{abstract}

6-DoF grasp detection of small-scale grasps is crucial for robots to perform specific tasks. This paper focuses on enhancing the recognition capability of small-scale grasping, aiming to improve the overall accuracy of grasping prediction results and the generalization ability of the network. We propose an enhanced receptive field method that includes a multi-radii cylinder grouping module and a passive attention module. This method enhances the receptive field area within the graspable space and strengthens the learning of graspable features. Additionally, we design a graspable balance sampling module based on a 3D segmentation network, which enables the network to focus on features of small objects, thereby improving the recognition capability of small-scale grasping. Our network achieves state-of-the-art performance on the GraspNet-1Billion dataset, with an overall improvement of approximately 10\% in average precision@k (AP). Furthermore, we deployed our grasp detection model on pybullet grasping platform and in real-world scenarios, which validates the effectiveness of our method.

\end{abstract}

\section{INTRODUCTION}

In recent years, grasp tasks for robotic arms have attracted significant attention in the fields of computer vision and deep learning \cite{du2021vision}. In the execution of grasping tasks by robots, grasp detection serves as a fundamental task, providing the robot with perceptual capabilities for the scene. The goal of grasp detection is, given a scene containing objects, to identify a set of grasp configurations (including grasp point locations, joint poses, etc.) where the robotic hand, when closed at that configuration, can robustly grasp the corresponding object. Traditional grasp detection methods are primarily model-based \cite{zhang2022robotic}. These methods generate multiple grasp poses satisfying stability conditions based on the 3D model of the object. Subsequently, they calculate grasp poses satisfying stability conditions after coordinate transformations based on the actual pose of objects in the scene, thereby completing the grasping process. However, this approach heavily relies on the accuracy of object pose estimation in the scene. With the advancement of deep learning technology and the reduction in the cost of depth-sensing devices like Kinect and RealSense, data-driven grasp detection methods leveraging deep learning have steadily gained popularity \cite{10149823}.
\begin{figure}
    \centering
    \includegraphics[width=1\linewidth]{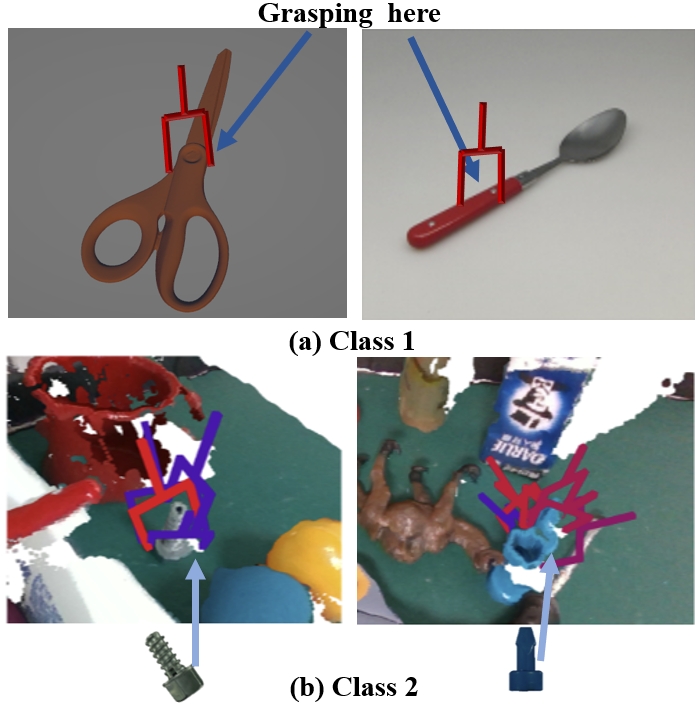}
    \caption{We categorize small-scale grasping into two classes. The first class involves grasping small parts of medium to large objects on the tabletop. The second class pertains to grasping small objects at the tabletop level.
}
    \label{fig1}
\end{figure}

Deep learning-based grasping can be categorized into planar grasping and 6-Degree-of-Freedom (6-DoF) grasping \cite{du2021vision}. Research on planar grasping detection primarily utilizes RGB or depth maps (RGB-D) as input and predicts a set of rotated bounding boxes to represent the grasping poses \cite{chu2018real}, \cite{asif2019densely}, \cite{jiang2011efficient}, \cite{mahler2017dex}, \cite{morrison2018closing}. Due to the significant limitation of planar grasping, which confines the grasping poses to vertical motions from top to bottom, this method faces challenges when operating in complex real-world scenarios. 6-DoF grasping, designed for more versatile scenarios, offers greater flexibility compared to planar grasping. 6-DoF grasping detection networks leveraging deep learning technologies can predict the 6 degrees of freedom of the grasping pose, along with the opening width of the two-finger gripper. Earlier methods often employed a two-step sampling and evaluation approach, such as GPD \cite{ten2017grasp} and PointnetGPD \cite{liang2019pointnetgpd}. Due to the typically low quality of sampled results, a large number of samples need to be evaluated, leading to considerable time consumption. With the development of grasping detection datasets, end-to-end networks are easier to design and can fully exploit the information inherent in the data. Fang et al. \cite{fang2020graspnet} proposed the GraspNet-1Billion grasping detection dataset, which has spurred various 6-DoF grasping detection research \cite{gou2021rgb}, \cite{li2021simultaneous}, \cite{wang2021graspness}, \cite{ma2023towards}. However, previous methods still have deficiencies in detecting small-scale grasps, Fig.\ref{fig1} illustrates the concept of small-scale grasping. The uneven distribution of grasp pose samples across different scales in the dataset leads to the deteriorated recognition performance for small-scale grasps \cite{ma2023towards}, thereby affecting the overall performance of the detection network. Although GSNet \cite{wang2021graspness} annotates the feasible grasp space, directly sampling in this space may easily overlook features beneficial for regressing small-scale grasps. Similarly, subsequent parts using single-scale cylinder grouping may encounter similar issues. This issue hinders the smooth execution of task-oriented grasping operations by robots. 

In this paper, we present a novel 6-DoF grasp detection network. Firstly, we propose enhanced receptive field method, which use Multi-radii Cylinder Grouping (MrCG) module to increase the receptive field area in the graspable space, and use Passive Attention (PA) module to enhance the sampled features. This enhancement facilitates better perception of fine details in small and medium-to-large objects. The feature based on graspable points also contributes to improved guidance for grasp pose prediction. Secondly, we introduce an 3D segmentation network based graspable balance sampling module. This module utilizes a pre-trained point cloud segmentation network to obtain the category of each point. Subsequently, it performs balanced sampling of graspable points learned by the network on each object, ensuring an equal number of sampled points for each object, therefore, small objects receive adequate attention from the model. Leveraging the 3D segmentation network, this method provides a solution for more advanced grasp tasks. Our approach outperforms previous state-of-the-art methods by 10\% in AP on the GraspNet-1Billion dataset. Furthermore, in terms of the evaluation benchmark based on grasp scale, our method surpasses other approaches. Finally, we construct a 6-DoF grasp platform using pybullet tools and real-world robotic arm to conduct grasp tests on our detection network. Experimental results demonstrate the precision of our method in accomplishing 6-DoF grasp tasks in clutter scenes. Our contribution can be summarized as follows:
\begin{itemize}
\item[(1)] 
We propose a enhanced receptive field method, which increases the model's receptive field area, enhancing the perception of small-scale graspable features.
\item[(2)] 
We propose an 3D segmentation network based graspable balance sampling method. This method improves the grasp detection network's perception of fine details in small and medium-to-large objects. Leveraging an segmentation network, this method provides a solution for more advanced grasp tasks.
\item[(3)] 
We validate our 6-DoF grasp detection method on the pybullet platform and real-word system. The experiments demonstrate that our method can accurately perform 6-DoF grasp tasks in clutter scenes.
\end{itemize}

\section{RELATED WORK}

Our focus is primarily on reviewing 6-DoF grasp detection methods based on deep learning. These methods can be broadly categorized into two types: the first category involves sampling and evaluation-based methods, while the second category comprises end-to-end network methods.

\textbf{Sampling and evaluation-based methods.} These methods initially generate multiple grasping poses for a given scene. Subsequently, evaluate the sample according to a quality estimation function, fulfilled by a deep neural network \cite{varley2015generating}, \cite{kappler2015leveraging},
 \cite{gualtieri2016high}, \cite{ten2017grasp}, \cite{liang2019pointnetgpd}. Some methods further employ optimization-based approaches to refine the samples and generate higher-quality grasp poses \cite{zhou20176dof}, \cite{yan2018learning}, \cite{mousavian20196}, \cite{murali20206}. A major drawback of sampling and evaluation methods is the need to strike a balance between computation time and the quantity of generated grasp poses. As a result, these methods typically require several seconds to run and can only generate dozens of grasp poses for a single scene. With the development of grasp detection datasets, the advantages of end-to-end networks have gradually become evident. The increased availability of data allows for leveraging the power of representation learning of end-to-end networks, which are easy to design, effectively utilize the information present in the data itself, and exhibit fast inference speeds.

\textbf{End-to-end network methods.} These methods can directly regress grasping poses on the scene. Early literatures use RGB-D inputs to generate grasp poses \cite{gou2021rgb}, \cite{lundell2021ddgc}, \cite{chen2023keypoint}. Fang et al. \cite{fang2020graspnet} introduced a large-scale 6-DoF grasp dataset and proposed an inference network based on point cloud deep learning to regress dense grasp poses on the scene. Wang et al. \cite{wang2021graspness} presented an evaluation method based on ``graspness'' to learn graspable regions. Ma et al. \cite{ma2023towards} investigated the issue of grasp scale imbalance in the GraspNet-1Billion dataset, identifying problems with the annotated grasp scales and proposing solutions. Breyer et al. \cite{breyer2021volumetric} introduced a Volume Grasping Network (VGN), which takes the Truncated Signed Distance Function (TSDF) representation of the scene as input and outputs predicted grasp quality, fixture direction, and opening width for each voxel in the queried 3D volume. Dai et al. \cite{dai2023graspnerf} proposed a 6-DoF grasp detection network, GraspNeRF, based on multi-view RGB inputs, addressing the problem of 6-DoF grasp detection for transparent and reflective objects.

However, grasp detection still faces challenges in terms of generalization and precision. Additionally, the detection capabilities for fine details in medium-to-large objects and small objects remain limited, leading to poor grasp pose availability. These challenges impede task-oriented grasping research.
\begin{figure*}
    \centering
    \includegraphics[width=1\linewidth]{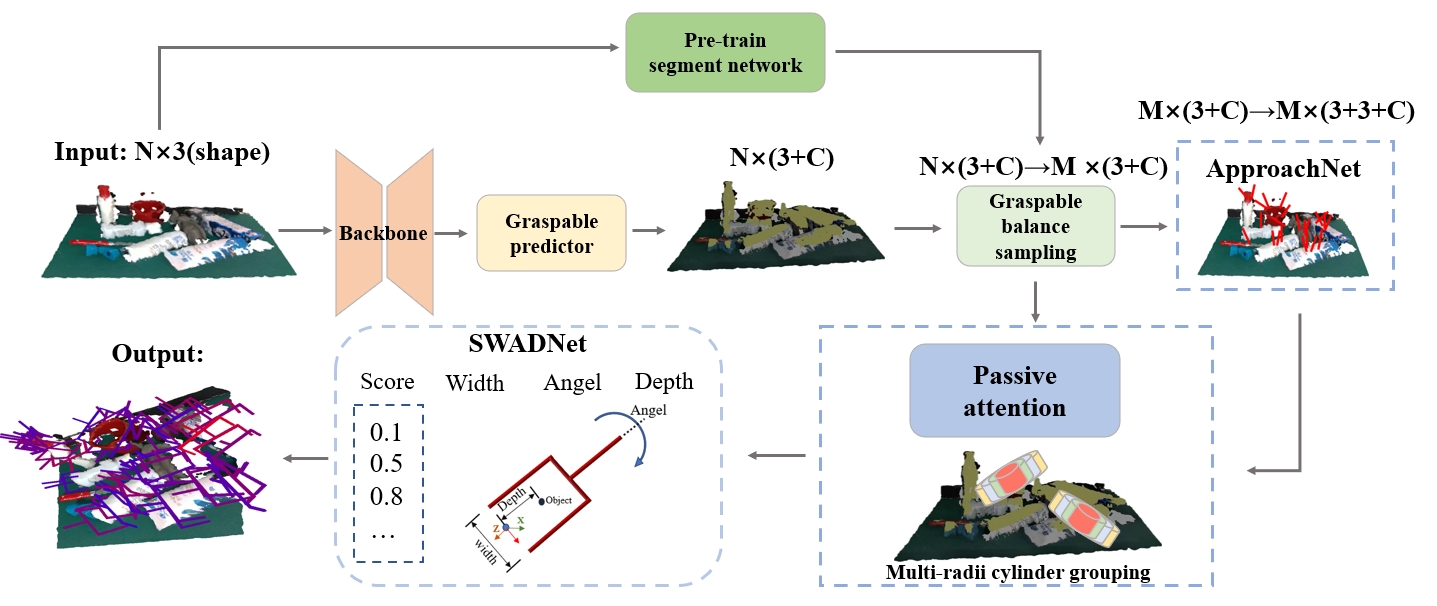}
    \caption{Pipeline. The network initially generates multiple features through the backbone, followed by the graspable predictor predicting points with high graspness. The graspable balance sampling module has two modes: during training, it directly uses the farthest point sampling without employing the guidance of pre-trained segmentation model features. The model-guided sampling is utilized only during inference. The features are then fed into the ApproachNet to select the optimal grasp views, which is subsequently input into our enhanced receptive field for cylinder grouping. Finally, the input is processed by SWADNet to output dense grasp poses.}
    \label{fig2}
\end{figure*}
\section{METHOD}
In this section, we first briefly introduce the pipeline overview of grasp detection in clutter with enhanced receptive field and graspable balance sampling, as shown in the Fig.\ref{fig2}. Then, we will focus on introducing our enhanced receptive field method and discuss two constituent modules: multi-radii cylinder grouping module and the passive attention module. Finally, we will present the graspable balance sampling module based on a pre-trained segmentation network, which exhibits two different forms during training and inference.

\subsection{Pipeline Overview}
\begin{figure}
    \centering
    \includegraphics[width=0.8\linewidth]{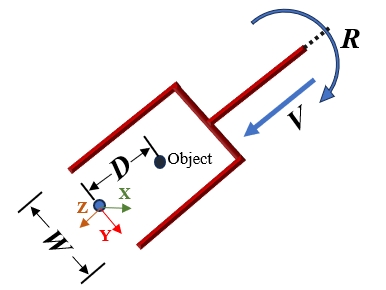}
    \caption{Grasp representation and gripper coordinate system.}
    \label{fig3}
\end{figure}

Our grasp detection network drew inspiration from \cite{wang2021graspness}, where they framed the grasping problem as a Bayesian problem involving the localization of grasp points and how to grasp, \textit{X}, \textit{Y}, \textit{Z} represent the grasp coordinates, \textit{V} represents the approach vector, \textit{R} represents the grasp rotation angle, \textit{D} represents the grasp depth, and W represents the width of the gripper opening. which is illustrated in Fig.\ref{fig3}. \textit{X}, \textit{Y}, \textit{Z}, and \textit{V} represent where to grasp, while \textit{R}, \textit{D}, and \textit{W} represent how to grasp, as illustrated in the formula:
\begin{equation}
P(Grasp) =P(R,D,W|X,Y,Z,V)P(X,Y,Z,V)
\end{equation}
We believe that learning based on graspable regions can enhance the accuracy of robotic grasp recognition. Initially, it is necessary to annotate spatial-level grasp ability in scenes of the dataset. Here, we briefly introduce the point-wise graspness score denoted as $\tilde{s}_i^{\mathrm{P}}$. This score involves sampling multiple grasp poses based on points in the scene, followed by an evaluation using force analysis conditions \cite{nguyen1988constructing}, \cite{mahler2017dex} and scoring as $q_{k}^{i,j}$, and $\textbf{1}(\cdot)$ is used to predict whether any grasp pose in space is valid. Grasps with collisions are filtered out. Indices \textit{i} and \textit{j} represent the point and approach vector, respectively. Index \textit{k} represents the grasp candidate $\mathcal{G}_{i,j}$, \textit{L} denotes the number of grasp candidates, \textit{c} is a scoring threshold, and $c_k^{i,j}$ represents the collision label for grasps, \textit{V} represents the numbers of view, the simplified version of the formula is as follows:
\begin{equation}
\begin{aligned}
\tilde{s}_i^p &= \frac{\sum_{j=1}^V \sum_{k=1}^L \mathbf{1}(q_k^{i,j} > c) \cdot \mathbf{1}(c_k^{i,j})}{\sum_{j=1}^V |\mathcal{G}_{i,j}|}, \quad i=1, \ldots, N
\end{aligned}
\end{equation}

Our innovation lies in proposing two methods to enhance the perceptual capabilities of small-scale grasping in the graspable space, as Fig.\ref{fig2} shows. Firstly, during the training phase of the network, we modify the cylinder grouping by using multiple cylinders to sample features for a single graspable point. We believe this method can yield a more powerful receptive field, reinforcing the accuracy of graspable point prediction in the earlier stages. 
Secondly, in the pure inference phase, we utilize a pre-trained point cloud segmentation network to guide the scene-level graspable point sampling, ensuring a balanced collection of graspable points on objects. Finally, our network can estimate denser grasp poses on the scene, where information about small-scale grasps may provide clues for future higher-level operations. Below, we will focus on elaborating the details of our methods.

\subsection{Enhanced Receptive Field}
\begin{figure}
    \centering
    \includegraphics[width=1\linewidth]{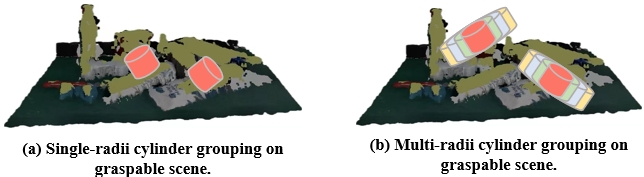}
    \caption{Schematic diagram of the cylinder grouping module. On the left is the conventional single-radii module, and on the right is our multi-radii module.}
    \label{fig4}
\end{figure}
As shown in the Fig.\ref{fig4}, previous works often used refined sampling points as the centers when performing cylinder grouping, setting a fixed radii for cylinder grouping \cite{fang2020graspnet}, \cite{wang2021graspness}. The consequence of this approach is a severe limitation on the receptive field area, leading to the failure of deep neural networks to learn the features of small parts of medium-to-large objects and grasp points of small-sized objects in clutter. To enhance the predictive capability for small-scale grasps and improve the network's regression performance, the network needs to consider more detailed geometric information. In this paper, we utilize the cylinder grouping method and enhance the receptive field after grasp sampling. as shown in the formula:
\begin{equation}
\mathcal{C}_{q}=\{{\textbf{v}_{{i}{{j}}}}|\|p_{i{j}}-p_{{i{j}}_{k}}\|\leq{r}_{q}\} , \quad q=1, \ldots, 4.
\end{equation}
Where ${\mathcal{C}}_{q}$ represents the result of a single cylinder grouping. This method clusters by limiting the radii $r_{q}$. The key difference from previous work lies in our use of four different radii. Meanwhile, the radii of these four sampled cylinders uniformly increase within the maximum width of the gripper of the manipulator. Finally, the multi-radii clustering groups are processed with :
\begin{equation}
\mathcal{C}=Concat\{\mathcal{C}_{1},\mathcal{C}_{2},\mathcal{C}_{3},\mathcal{C}_{4}\}.
\end{equation}

\begin{figure}
    \centering
    \includegraphics[width=1\linewidth]{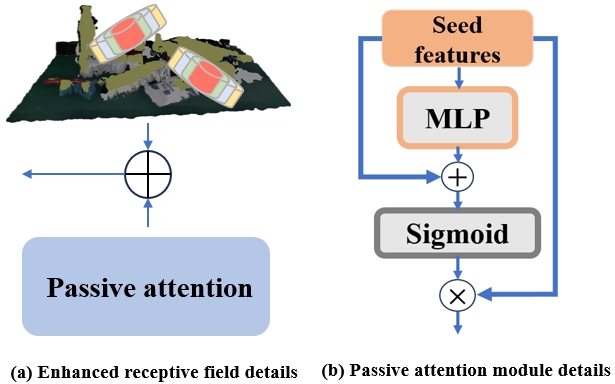}
    \caption{On the left is our designed enhanced receptive field method. On the right is the pipeline of our passive attention module.}
    \label{fig5}
\end{figure}

Simultaneously, this module requires a passive attention module to guide the fused features, as shown in the Fig.\ref{fig5}. This module is trained using features sampled from the grasp scene. The reason for incorporating this module may be the introduction of noise due to the increased receptive field. We believe that performing multi-scale grouping on the basis of graspable scenes can strengthen the predictive capabilities of the grasp network and improve its generalization performance. Although an increased receptive field may lead to noise perception, the robustness of the network against interference is enhanced in graspable scenes, as the graspness score in these scenes is continuous. This approach deepens the network's understanding of scene details, allowing for a comprehensive perception of geometric features in small parts.

\subsection{Graspable Balance Sampling}

\begin{table*}
\caption{Ablation study of the proposed modules results on RealSense D435}
\centering

\begin{tabular}{c|ccc|ccc|ccc}
\hline
\multirow{2}{*}{\textbf{Model}} &  & \textbf{Seen} &    &    & \textbf{Similar} &    &    & \textbf{Novel} &    \\ 
\cline{2-10} 

                  &\textbf{ AP }      & $\textbf{AP}_{0.8}$  & $\textbf{AP}_{0.4}$ & \textbf{AP} & $\textbf{AP}_{0.8}$      & $\textbf{AP}_{0.4}$ & \textbf{AP} & $\textbf{AP}_{0.8}$    & $\textbf{AP}_{0.4}$ \\ 
                  \hline
w/o PA       &73.46      & 85.13    & 68.33 & 64.21  & 77.02 & 56.38   & 26.63   & 33.19  & 14.04   \\
w/o MrCG       &69.92      & 81.51    & 64.27 & 60.99  & 73.93  & 52.22 & 25.70   &32.13   & 13.67   \\
Ours &74.33     & 85.77    & 63.89 & 64.36  & 76.76  & 55.25  & 27.56 & 34.09 &20.23    \\
 \hline
\end{tabular}
 \label{tab1}
\end{table*}

Previous research utilized a point cloud segmentation network to fully leverage the capabilities of farthest point sampling (FPS), ensuring an equal number of points are sampled on each object \cite{ma2023towards}. However, they did not conduct sampling under conditions that reasonably position the grasping space. This approach fails to guarantee that the sampled points do not contain excessive noise, which may hinder the overall grasp prediction performance, resulting in the prediction of low-quality grasps.

\begin{algorithm}
    \SetKwInOut{Input}{Input}
    \SetKwInOut{Output}{Output}

    \Input{N points}
    \Output{M points}
    \BlankLine


   $N_{ps}\leftarrow M/idx$
   
   \For{$j\leftarrow 1$ \KwTo idx}
    {
       \uIf
           {
            $N_{g} = 0$
           }
            {
             Use FPS on the object points;      
                 }
       \uElseIf
               {
                     $N_{g} < N_{ps}$
               }
                {
                  Sample all graspable points on the object, then supplement with FPS other points on the object;
                }
       \Else{
              Use FPS on the object graspable points;
             }

}

    \caption{Balanced Sampling in the Graspable Space}
    
\end{algorithm}

To address this issue, we propose an 3D segmentation network based graspable sampling method, which use GBS module for inference. During the grasp pose inference phase, we utilize a pre-trained lightweight point cloud segmentation network \cite{xie2021unseen}. Due to segmentation, we can effectively identify the class ownership of the point cloud in the scene, enabling us to commence balanced sampling. The details of our algorithm are described in Algorithm 1. Point clouds are acquired from both graspable scenes and original scenes. Our goal is to sample more graspable points as inputs for our network. However, on certain objects, it may be challenging to predict enough graspable points. Firstly, The number of grasp points needed for each object is calculated. Then, we set up three sampling scenarios. In the first scenario, when there are no graspable points on the object, we perform FPS on the original point cloud in the scene. In the second scenario, if the number of graspable points on the object is less than the required sampling points, all graspable points on the object are sampled first, and then FPS is performed on the remaining points. In the third scenario, when the number of graspable points on the object meets the sampling point requirement, FPS is directly performed on the graspable points of the object. Simultaneously, our method can more efficiently regress semantic-based grasp poses in everyday usage scenarios. This approach provides rich visual clues for more advanced task operations.

\section{EXPERIMENTS}

\subsection{Implementation Details}

\textbf{Benchmark dataset and metric.} Our network is trained and tested on the GraspNet-1Billion dataset \cite{fang2020graspnet}, which includes 190 scenes. Each scene contains information captured from 256 viewpoints, and dense grasp poses are annotated for each scene. The test set is divided into three categories based on difficulty: seen, similar, and novel. We employ two evaluation methods to assess our network. Firstly, we use precision@k as our evaluation metric, which measures the precision of the top k-ranked grasps. \textbf{AP}µ represents the average precision@k under a given friction coefficient µ. This evaluation method utilizes dynamic force closure analysis, which aligns better with real-world grasp success conditions. Secondly, we aim to validate that our network contributes to improving the recognition performance of small-scale grasps. We adopt the method proposed in \cite{ma2023towards}, which defines the grasp scale as the gripper's opening width and categorizes it into three classes: widths in 0cm-4cm, 4cm-7cm, and 7cm-10cm as small-scale, medium-scale, and large-scale. Following the same dynamic force closure analysis, we evaluate the scene's $\textbf{AP}_S$, $\textbf{AP}_M$, and $\textbf{AP}_L$.

\textbf{Network Implementations}. Our network implementations involve the use of 4D convolutions \cite{choy20194d} in the backbone to extract and learn features. The output features are increased to 512 channels. The graspable predictor in our network employs MLP for predicting the graspable positions. The shape of the MLP for learning grasp points is (512, 3), and for learning viewpoints, it is (512, 3). In the training phase for grasp point sampling, FPS is directly used in the graspable space, while in the inference phase, our segmentation network based balance graspable sampling method is employed for sampling. In the enhanced receptive field stage, we perform multi-scale cylinder sampling using the sampled points and features, with four different radii (0.0125m, 0.025m, 0.0375m, 0.05m), then concatenate the features of the four groups and pass them through a MLP with a shape of (1024, 512). In the PA module, the shape of the MLP is (512, 512). SWADNet is used for yielding final grasp rotation angles and grasp depth, and the shapes of the two MLPs inside it are (512, 256) and (256, 48).

\textbf{Training and Inference}. Our model is implemented with PyTorch and trained on one NVIDIA Tesla V100 GPU for 13 epochs with Adam optimizer \cite{kingma2014adam} and the batch size of 4. The learning rate is 0.001 at the ﬁrst epoch, and multiplied by 0.95 every one epoch. The network takes about 2 day to converge. In inference with collision detection, we only use one GPU for prediction.

\subsection{Ablation Study}

\begin{table}
\centering
\caption{Ablation study of the proposed modules in small scale grasping results on RealSense D435}
\begin{tabular}{c|ccc}
\hline
\textbf{Model} & \multicolumn{1}{l}{\textbf{Seen}} & \multicolumn{1}{l}{\textbf{Similar}} & \multicolumn{1}{l}{\textbf{Novel}} \\ \hline
GSNet      & 20.07          &6.75 & 9.59           \\
Ma et al.        & 18.29          & \textbf{10.03}         & 9.29           \\ \hline
w/o PA    & 22.11          & 8.82          & 11.36          \\
w/o MrCG   & 21.22          & 7.64          & 10.76          \\
Ours       & 23.01          & 8.89          & 11.33          \\
Ours + GBS & \textbf{23.67} & 9.21          & \textbf{11.38} \\ \hline
\end{tabular}
\label{tab2}
\end{table}

\begin{table*}
\centering
\caption{Ablation study in full scale grasping results on RealSense D435}
\begin{tabular}{c|cccc|cccc|cccc}
\hline
\multirow{2}{*}{\textbf{Model}} &
  \multicolumn{4}{c|}{\textbf{Seen}} &
  \multicolumn{4}{c|}{\textbf{Similar}} &
  \multicolumn{4}{c}{\textbf{Novel}} \\ \cline{2-13} 
 &
  $\textbf{AP}_{S}$ &
  $\textbf{AP}_{M}$ &
  \multicolumn{1}{c|}{$\textbf{AP}_{L}$} &
  \textbf{Mean} &
  $\textbf{AP}_{S}$ &
  $\textbf{AP}_{M}$ &
  \multicolumn{1}{c|}{$\textbf{AP}_{L}$} &
  \textbf{Mean} &
  $\textbf{AP}_{S}$ &
  $\textbf{AP}_{M}$ &
  \multicolumn{1}{c|}{$\textbf{AP}_{L}$} &
  \textbf{Mean} \\ \hline
GSNet &
  20.07 &
  65.11 &
  \multicolumn{1}{c|}{72.41} &
  52.53 &
  6.75 &
  50.51 &
  \multicolumn{1}{c|}{64.72} &
  40.66 &
  9.59 &
  24.20 &
  \multicolumn{1}{c|}{26.25} &
  20.01 \\
Ma et al.  &
  18.29 &
  52.6 &
  \multicolumn{1}{c|}{64.34} &
  45.08 &
  10.03 &
  42.77 &
  \multicolumn{1}{c|}{57.09} &
  36.63 &
  9.29 &
  18.74 &
  \multicolumn{1}{c|}{24.36} &
  17.46 \\ \hline
Ours &
  23.01 &
  67.67 &
  \multicolumn{1}{c|}{76.95} &
  55.88 &
  8.89 &
  53.88 &
  \multicolumn{1}{c|}{67.16} &
  43.31 &
  11.33 &
  25.58 &
  \multicolumn{1}{c|}{27.44} &
  21.45 \\
Ours + GBS &
  23.67 &
  67.54 &
  \multicolumn{1}{c|}{78.53} &
  \textbf{56.58} &
  9.21 &
  54.39 &
  \multicolumn{1}{c|}{68.83} &
  \textbf{44.14} &
  11.38 &
  26.17 &
  \multicolumn{1}{c|}{28.20} &
  \textbf{21.92} \\ \hline
\end{tabular}
\label{tab3}
\end{table*}

\begin{table*}
\caption{GraspNet-1Billion evaluation results on RealSense D435}
\centering

\begin{tabular}{c|ccc|ccc|ccc}
\hline
\multirow{2}{*}{\textbf{Model}} &  & \textbf{Seen} &    &    & \textbf{Similar} &    &    & \textbf{Novel} &    \\ 
\cline{2-10} 

                  &\textbf{ AP }      & $\textbf{AP}_{0.8}$  & $\textbf{AP}_{0.4}$ & \textbf{AP} & $\textbf{AP}_{0.8}$      & $\textbf{AP}_{0.4}$ & \textbf{AP} & $\textbf{AP}_{0.8}$    & $\textbf{AP}_{0.4}$ \\ 
                  \hline
GPD              &22.87      & 28.53    & 12.84 & 21.33  & 27.83 & 9.64   & 8.24   & 8.89  & 2.67   \\
PointnetGPD       &25.96      & 33.01    & 15.37 & 22.68  & 29.15  & 10.76 & 9.23   & 9.89  & 2.74   \\
GraspNet-baseline &27.56     & 33.43    & 16.95 & 26.11  & 34.18  & 14.23  & 10.55 & 11.25 &3.98    \\
Gou et al.         &27.98    & 33.47 &17.75    &27.23    & 36.34  &15.60   & 12.25 & 12.45 &5.62    \\
GSNet             &67.12    & 78.46 & 60.90   &54.81    & 66.72  & 46.17   &24.31 & 30.52  & 14.23   \\
Ma et al.          & 63.83   & 74.25  & 58.66  &58.46    & 70.05  & 51.32  & 24.63 & 31.05 &12.85\\ \hline
Ours       &\textbf{74.33}     & \textbf{85.77}    & \textbf{63.89} & \textbf{64.36}  & \textbf{76.76}  & \textbf{55.25}  & \textbf{27.56} & \textbf{34.09} &\textbf{20.23}  \\
\hline

\end{tabular}
\label{tab4}
\end{table*}


In this section, we primarily validate the effectiveness of our proposed MrCG and PA modules on the GraspNet-1Billion dataset \cite{fang2020graspnet}. Initially, we employed the benchmarks \cite{fang2020graspnet} to evaluate these modules. As shown in the Table \ref{tab1}, the MrCG module achieved a significant improvement in most evaluation metrics. Moreover, when MrCG and PA module were used in combination, there was a notable enhancement in the most stringent evaluation metric, $\textbf{AP}_{0.4}$ by novel. This experiment demonstrates that our approach comprehensively improves grasp detection accuracy and exhibits certain generalization performance improvements compared to previous methods.

Secondly, our approach focuses on enhancing the recognition capability for small-scale grasps. We utilized \cite{ma2023towards} to propose an evaluation standard tailored to grasp scales to assess the model's ability to recognize small-scale grasps. We evaluated the effectiveness of the PA and MrCG modules in improving small-scale grasp recognition. Simultaneously, we assessed the effectiveness of using the GBS module for inference on small-scale grasps. The experiments indicate that there is a certain effect when using the GBS module for grasping. As shown in Table \ref{tab2}, the MrCG module contributes the most, and compared to previous methods, we achieved state-of-the-art performance in the seen objects and novel objects tests but slightly lower results in similar, possibly due to the use of cost-sensitive learning in \cite{ma2023towards}, \cite{elkan2001foundations} approach. They designed a special loss function for weighting the errors of samples at different grasp scales, based on the frequency of grasp scale categories in the dataset.

Although our method focuses on improving the detection capability of small-scale grasps, our approach outperforms previous methods in grasp detection across all scales, as shown in Table \ref{tab3}, this improvement is particularly significant when we employ GBS for grasp pose inference.

\subsection{Comparing with Representative Methods}

We conducted comparisons with representative methods on the GraspNet-1Billion dataset. Methods utilizing point cloud input generally outperform those using RGB images, and our focus was on comparing with methods which take point cloud as input. We followed the testing methodology outlined in \cite{fang2020graspnet}, conducting tests on three object categories. In our model, along with \cite{wang2021graspness}, and \cite{ma2023towards}, we utilized collision detection for final predictions to achieve better results. The results of our comparative evaluation are reported in Table \ref{tab4}, We present only the RealSense section of the dataset because previous work has shown that training with Kinect data under the same network architecture results in suboptimal inference performance \cite{fang2020graspnet}, whereas data collected using a RealSense camera yields better detection outcomes.

Our method achieved state-of-the-art performance compared to previous approaches. Using the FPS mothod for inference, our network showed an improvement of approximately 10\% in AP metrics compared to previous methods. Notably, in the most challenging novel scenarios, our network's performance improved by 9.6\% compared to previous methods.

\subsection{PyBullet Grasping Experiment}

A platform for testing the grasping capability of small objects was established based on the pybullet robot tools \cite{wang2023policy}. Here are the details of our implementation. The Franka Panda gripper was utilized for grasping, aiming to validate the algorithm's effectiveness while conserving computational resources. To simulate the working space of a real manipulator, we filtered out grasping poses with an angle exceeding a threshold with the world coordinate system's Z-axis \cite{fang2023anygrasp}. During the grasping process, the gripper was initially positioned to a pre-grasping pose, with the grasping pose set along the object's approach vector. When executing the grasp, the gripper moved along the approach vector. A successful grasp was defined as holding the object, moving to the pre-grasping position, and maintaining stability for a certain duration.

\begin{table}
\centering
\caption{test objects list in YCB dataset}
\begin{tabular}{|c|c|}
\hline
Objects categories & IDs                                                                                   \\ \hline
Small              & \begin{tabular}[c]{@{}c@{}}11, 12, 31, 37, 38, 72-i, \\ 72-d, 72-f, 73-c\end{tabular} \\ \hline
Medium to large    & \multicolumn{1}{l|}{4, 6, 10, 13, 16, 19, 21, 35}                                     \\ \hline
\end{tabular}
\label{tab5}
\end{table}

\begin{figure}
    \centering
    \includegraphics[width=1\linewidth]{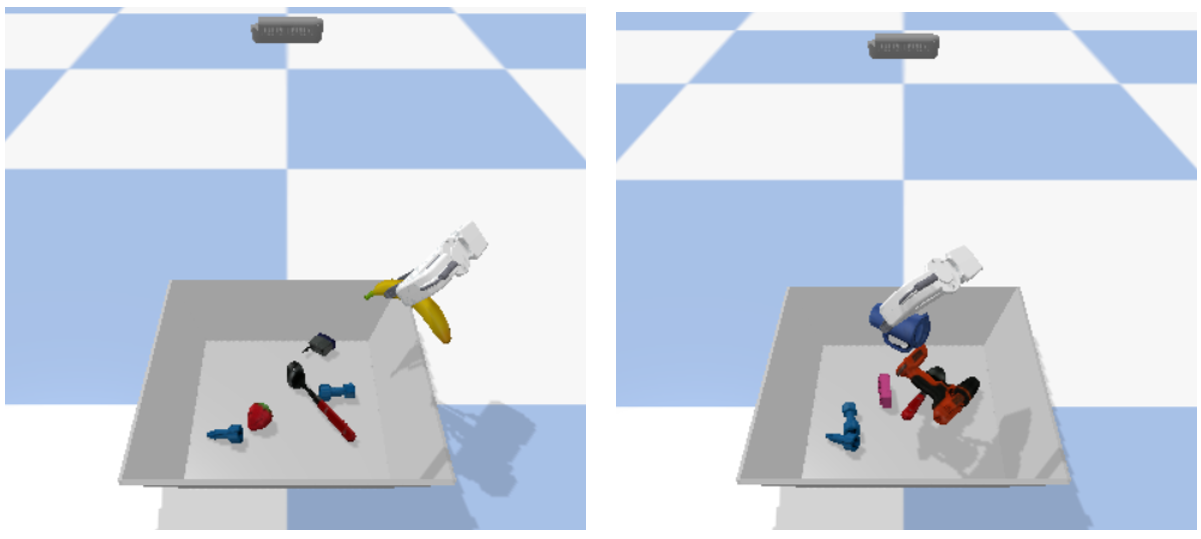}
    \caption{We divided the scenes into two categories: the left category is small scale objects scene, and the right category is mixed scale objects scene.}
    \label{fig6}
\end{figure}

\begin{table*}
\centering
\caption{pybullet grasping results in cluttered scenes}
\begin{tabular}{c|c|c}
\hline
\textbf{Model} & \textbf{Small Scale objects Scenes SR} & \textbf{Mixed Scale objects Scenes  SR} \\ \hline
GSNet\cite{wang2021graspness}           & 0.86                          & 0.84                        \\
Ours           & \textbf{0.95}                          & \textbf{0.87}                       \\ \hline
\end{tabular}
\label{tab6}
\end{table*}

17 models are selected from the YCB dataset \cite{calli2015ycb} and divided them into two categories, as shown in Table \ref{tab5}, one for objects suitable for small-scale grasping poses, and the other for objects suitable for medium to large-scale grasping poses. We designed two types of experimental scenes based on object classification. As shown in Fig.\ref{fig6}, the first type of scene contains only six small-scale objects. In the second type of scene, we randomly select six objects from the already chosen models to compose a new scene for grasping experiments. We refer to such scenes as mixed-scale objects scenes. It is worth mentioning that the objects in our scenes are randomly placed and cluttered, rather than being isolated. 

As shown in Table \ref{tab6}, we conducted comparative experiments with the state-of-the-art methods in our pybullet environment, and tested our proposed model in two types of scenes, The scene completion rate of all our grasping tests is 1. Firstly, In the small-scale objects scenes, ours model attempted 63 grasps in total, achieving successful grasps 60 times, resulting in success rate (SR) of 0.95. GSNet attempted 70 grasps in total, achieving successful grasps 60 times, resulting in SR of 0.86. Our model gains an almost 10\% improvement compared to GSNet \cite{wang2021graspness}. Secondly, for mixed-scale objects scenes, we attempted 107 grasps in total, achieving successful grasps 90 times, resulting in SR of 0.87. GSNet \cite{wang2021graspness} attempted 107 grasps in total, achieving successful grasps 90 times, resulting in success rate (SR) of 0.84. This experimental results validates that our method enhances the recognition capability of deep learning networks for small-scale grasping in clutter scenes, while also effectively addressing medium to large scale grasping recognition. Compared to previous methods, our model has achieved notable improvements in understanding objects grasping.

\subsection{Real-world Grasping Experiment}

\begin{figure}
    \centering
    \includegraphics[width=0.9\linewidth]{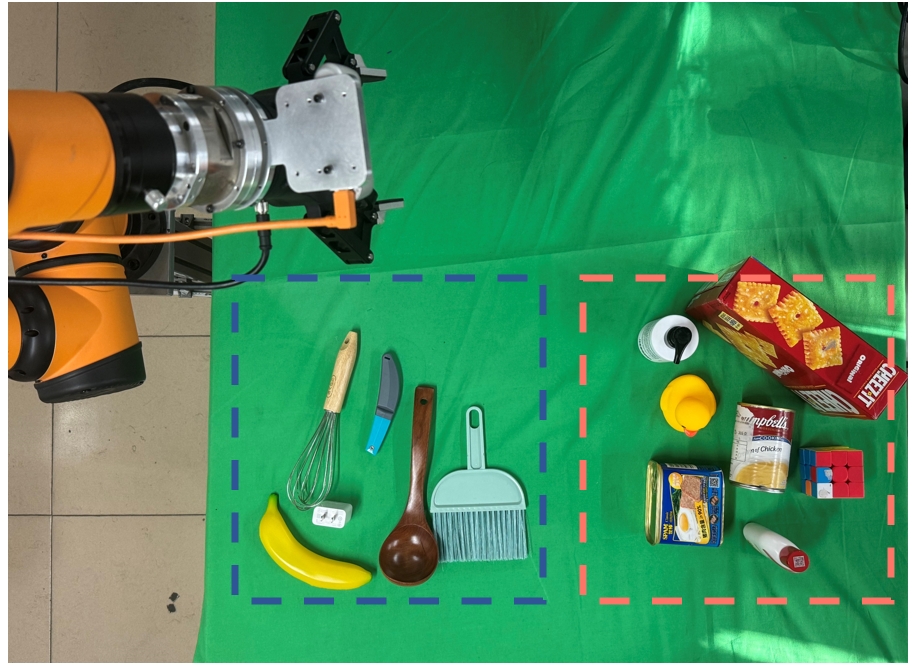}
    \caption{Robot and object settings in real-world experiment: the blue box is used for testing small-scale grasps, while the red box contains objects for testing medium to large-scale grasps.}
    \label{fig7}
\end{figure}
To verify the effectiveness of our grasp detection method in the real-world, we set up a physical scenario for grasping experiments. As shown in Fig.\ref{fig7}, our grasping platform consists of an AUBO-i5 collaborative robot, a RealSense camera, a 2-Fingered Modular Changing Hand, and two categories of objects to be grasped. The grasping strategy and object deployment are similar to the simulation experiments, mainly testing our method's capability in small-scale and full-scale grasping. It is worth mentioning that in the experiments testing small-scale grasping ability, we selected some tools frequently used by humans in daily life. Evaluating our method on these objects is crucial because it needs to provide a foundation for generalized robotic manipulations in the future.

\begin{figure}
    \centering
    \includegraphics[width=1\linewidth]{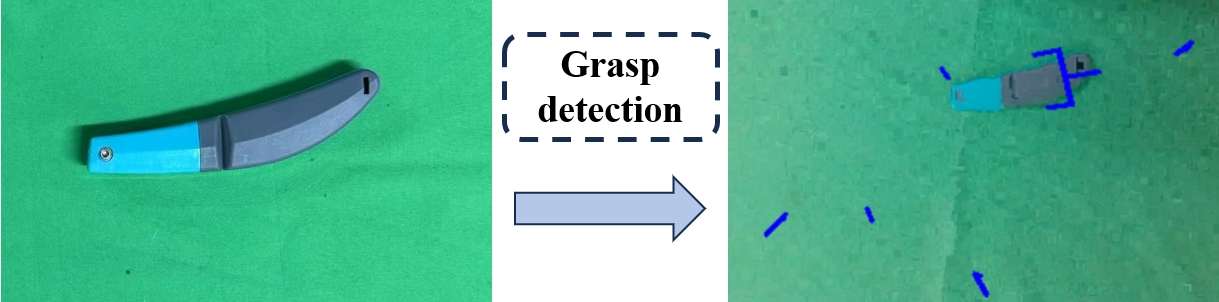}
    \caption{Failure cases in real-word experiment}
    \label{fig8}
\end{figure}

Our SR is 0.83 for small-scale grasping and 0.91 for mixed-scale grasping. This result is close to our success rate in simulation because the training datasets \cite{fang2020graspnet} was collected from real-world. However, it is challenging to capture the complete depth for flat objects. As shown in the Fig.\ref{fig8}, when the grasp detection performance is less than optimal, our experimental method tends to choose some noisy poses. In such cases, using the highest grasp score is not suitable. One possible solution is to use image based instance segmentation methods to filter the grasp poses.

\section{CONCLUSIONS}

In this paper, we propose an enhanced receptive field method, which increases the model's receptive field based on graspable space. This modification allows the model to pay more attention to subtle features on objects. Additionally, we propose a segmentation network based graspable balance sampling method. This method utilizes a point cloud segmentation model to extract points belonging to the grasped object, effectively filtering out noise. Graspable points on each object are uniformly sampled, ensuring that features of small objects are not overlooked and enhancing the recognition capability for small-scale grasping. The segmentation model also provides a solution for semantic grasping. We conducted extensive experiments to evaluate our proposed methods. The accuracy tests of the grasping detection network demonstrate that, compared to previous methods, our method improves the recognition ability for small-scale grasping at the visual level, enhancing overall network generalization and accuracy. Furthermore, objects grasping experiments confirm that our grasping detection network can predict effective grasping poses. In the future, our work can provide grasp perception capabilities for task-oriented robotic manipulation.





\section*{ACKNOWLEDGMENT}
The research was partially supported by the National Natural Science Foundation of China (No. 52005120) and the Interdisciplinary Team of Intelligent Elderly Care and Rehabilitation in the ``Double First-class'' Construction of Beijing University of Posts and Telecommunications in 2023 (No. 2023SYLTD04).



\end{document}